# Pearmut: Human Evaluation of Translation Made Trivial

**Vilém Zouhar**
ETH Zurich

**Tom Kocmi**
Cohere

## Abstract

Human evaluation is the gold standard for multilingual NLP, but is often skipped in practice and substituted with automatic metrics because it is notoriously complex and slow to set up with existing tools with substantial engineering and operational overhead. We introduce Pearmut, a lightweight yet feature-rich platform that makes end-to-end human evaluation as easy to run as automatic evaluation. Pearmut removes common entry barriers and provides support for evaluating multilingual tasks, with a particular focus on machine translation. The platform implements standard evaluation protocols, including DA, ESA, and MQM, and is extensible to support new protocols. It features document-level context, absolute and contrastive evaluation, attention checks, ESA$^{\text{AI}}$ pre-annotations and both static and dynamic assignment strategies. Pearmut enables reliable human evaluation to become a practical, routine component of model development and diagnosis rather than an occasional effort.[1]

## 1 Introduction

While human annotations are the lighthouse for evaluating NLG outputs, they are commonly sidelined by automatic evaluation. And in their absence, it is possible to hillclimb automated metrics which then lead to false conclusions (Kocmi et al., 2025; Lavie et al., 2025). While cost and access to human annotators are part of the reason, limitations in usability of existing evaluation tools further discourage researchers from running even limited human evaluations.

As a result, **human evaluation in papers is scarce**. Through a review of 82 translation modelling papers at *ACL 2025 papers we found that 55 (68%) do not report any human evaluation (see Appendix G), still alarming finding as Marie et al.'s (2021) (>90% of 769 did not use human evaluation). This can be partly explained by the lack of access to multilingual evaluators. When human evaluation does take place, it is oftentimes with an ad-hoc non-standard annotation protocol that is poorly documented, which makes it irreproducible, unauditable and increases annotation noise, making the study inconclusive (Freitag et al., 2021). We attribute this to the high bar of setting up human evaluation framework and absence of a go to default.

We present **Pearmut**, a platform for evaluating and reviewing of multilingual tasks. Unlike general-purpose tools, Pearmut is engineered specifically for multilingual and translation evaluation. It supports best practices (Kocmi et al., 2025), including span-level error marking (ESA Kocmi et al., 2024a, MQM Freitag et al., 2021), pre-annotations (ESA$^{\text{AI}}$ Zouhar et al., 2025b), tutorial tasks, attention checks to ensure data quality, reviewing existing annotations, monitoring, and statistical analysis. Pearmut focuses on the following use-cases primarily for both textual and multimodal (audio, video, image) translation evaluation:

- rapid model prototyping,
- larger-scale evaluation and shared tasks,
- evaluation to support academic papers,
- benchmarking production systems.

The tool is intended for: academics and industry practitioners evaluating models, and shared task organizers (e.g. WMT, IWSLT). The annotators in these cases can be the practitioners themselves, their colleagues with required language skills, translation vendor companies, or crowd-workers.

In this paper, we describe the publicly available tool Pearmut (Section 3) and provide empirical evidence of its usefulness and ease-of-use (in comparison to existing tools) through two experimental case studies focused on researchers setting up an evaluation platform (Section 4), and annotators using the evaluation platform (Section 5).

[1] github.com/zouharvi/pearmut with MIT license

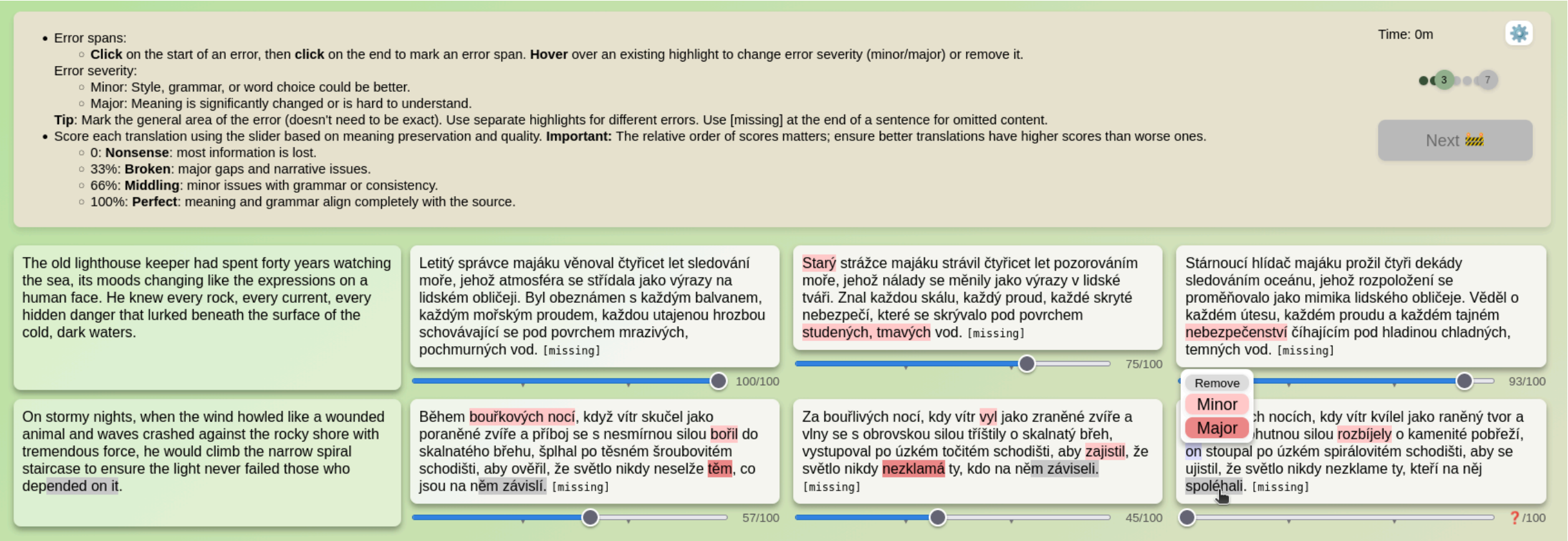

Figure 1: Pearmut annotation interface with document-level contrastive ESA protocol together with guidelines.

## 2 Related Work

**Human evaluation.** Automatic metrics aim to imitate human evaluation. Still, even if a practitioner decides to hire human annotators, it is unclear how the assessment should be solicited. This can range from a simple scale (Graham et al., 2015), ranking (Bojar et al., 2016), to marking specific errors (Freitag et al., 2021), or their combination (Kocmi et al., 2024a). The way the human judgment is solicited is referred to as *annotation protocol*, described in Section 3.

A large body of work is devoted to increasing the quality of annotations (Graham et al., 2013b; Graham et al., 2017; Khashabi et al., 2022) or making the evaluation process more efficient. For example, Zouhar et al. (2025b) use AI-prefilled error spans, turning the human judgment of translation quality into human post-editing of translation quality. On a higher level, Sakaguchi et al. (2014); Zouhar et al. (2025a); Saldías Fuentes et al. (2022) steer the focus on which evaluation items are human-annotated to avoid spending the annotation budget on uninformative items.

**Tools for human evaluation.** The landscape of evaluation *tools* is polarized. On one end stands Appraise (Federmann, 2018), a long-standing annotation platform behind the WMT shared tasks (Kocmi et al., 2025, inter alia). However, it is difficult to deploy or modify for experiments. ChatbotArena (Chiang et al., 2024) is a modern tool for pairwise comparisons, but specialized for crowdsourced evaluations based on free-form chat interactions, which is not fitting for many multilingual tasks.

On the other end are general-purpose tools[2] like Label Studio (Tkachenko et al., n.d.), Potato (Pei et al., 2022), or Factgenie (Kasner et al., 2024). While these offer excellent user experience for classification or span tagging, they lack the specific logic required for modern translation evaluations. Similarly, interactive prototyping frameworks such as Gradio (Abid et al., 2019) and Streamlit are popular for model demonstration but lack state management and infrastructure. Adapting these modern tools often necessitates involved configurations and custom backend engineering to support best-practice annotation protocols.

Pearmut occupies the middle ground: lightweight like Potato, but domain-specialized like Appraise. It also contains statistical analysis of results (Appendix A.3, two-sided t-test with $\alpha = 0.05$) to promote good and standardized scientific practices, missing in other tools. We consider the setup ease to be especially important to support good research practices regarding evaluations (similar to the effect of Post's (2018) SacreBLEU). See Appendix Table 3 for a comparison across evaluation tools.

## 3 Pearmut

Pearmut is a pip-installable package that aims to be minimal and provide reasonable defaults specifically for machine translation evaluation. Running a campaign consists of 3 commands: pip install, adding a campaign, and launching the server. See Appendix F for example campaign definition JSONs and Appendix A.1 for architecture details.

**Annotation protocols.** Annotation protocols and interfaces are often interchanged, but they are not the same. An interface provides the infrastructure and visual environment for annotators, whereas a protocol defines the rules and metrics, such as sliding scales, error spans or task distribution, used for the assessment. The interface is thus a specific implementation of an annotation protocol, which is more abstract. Pearmut supports a range of evaluation protocols, which differ in speed (cost), depth, and annotation quality:

- Direct Assessment (**DA**, Graham et al., 2015): Scale from 0 (poor) to 100 (perfect).
- Multidimensional Quality Metrics (**MQM**, Freitag et al., 2021): Error spans with severities (*minor*, *major*) and categories, such as *Accuracy/Overtranslated*.
- Error Span Annotation (**ESA**, Kocmi et al., 2024a): Error spans with severities but not categories, extended by DA as a final score.
- Pre-filled Error Span Annotation (**ESA$^{\mathbf{AI}}$**, Zouhar et al., 2025b): Same as ESA but the annotators start with spans pre-filled by a quality estimation metrics that they post-edit.

[2]We exclude MTurk and Prolific as their role is worker-aggregation rather than providing interfaces.

All of these protocols work on document-level (multiple sources shown together) which can be shown in contrastive manner, i.e. multiple outputs shown in parallel. This extends the pairwise side-by-side (Novikova et al., 2018; Sakaguchi and Van Durme, 2018; Akhbardeh et al., 2021; Song et al., 2025) for faster and more accurate assessment. See a document-level contrastive example in Figure 1.

**Assignment.** Assignment determines which evaluation items are shown to which users and in which order. This order is important because it affects the annotator behavior and their annotations (Mathur et al., 2017), and is critical to do efficiently especially when dealing with limited budgets or volunteer efforts. Annotators get access to the platform via pre-generated magic links but the exact item assignment depends on one of the three assignment strategies that is selected during the setup (also illustrated in Figure 5).

Most common, **task-based assignment** offers control over the annotation process by matching specific evaluation items to specific users beforehand. This is brittle if an annotator drops out or when the annotation times vary greatly and generally requires more management effort.

A simpler approach is to supply all evaluation items and have all annotators work on them simultaneously. **Single-stream assignment** maximizes throughput, though can lead to occasional double-annotations and there is less control on the sequence order, as the next evaluation items are assigned fully at random from the pool of unannotated ones. The randomness balances but does not guarantee annotations across different models and documents (e.g. a pessimistic annotator can see a particular model more often), which is key for unbiased evaluation.

To optimize limited budgets, Pearmut supports **dynamic assignment** (Kiela et al., 2021, inter alia). This strategy, implemented as an optional extensibility showcase, uses $\varepsilon$-greedy sampling to prioritize top-performing models, thereby accelerating finding the best models, compared to indiscriminate evaluation (see Appendix A.2).

**Dashboard.** When starting a campaign, Pearmut generates annotation magic links together with a link for a *dashboard* that can be used for:

- monitoring annotation progress, time, and attention check success rate,
- viewing existing annotations,
- redistributing annotation tasks, and
- viewing intermediate and exporting final model rankings with statistical analysis[3]

An example of a dashboard with two campaigns loaded is shown with details in Appendix A.3.

**Additional features.** Pearmut supports features, which are occasionally missing from other tools:

- Re-doing existing annotations without destructing the already-collected data
- Interactive tutorials and attention checks (e.g. Graham et al., 2013a); see Appendix A.4.
- Multimodal inputs and outputs; instead of texts, one can use audio, video, or other interactive HTML elements; see Figure 3.
- Post-editing and custom evaluation sliders (e.g. *Diversity* or *Appropriateness*) to accommodate non-MT tasks, such as source-only evaluation.
- Approximate alignment; to quickly orient between the source, the reference, and the translation, we show approximate positions on mouse hover, as in Figure 1 (user can turn off).
- Word/character-level error span annotations.
- Free-form comment box for issues or comments.

## 4 Case Study with Researchers

To validate the low setup complexity, we asked five NLP practitioners to set up an evaluation campaign using Pearmut, Appraise, Factgenie, Potato, and Label Studio (see details in Appendix B). We focused on (1) the time it took to set-up and annotate a few examples, and (2) self-reported ease-of-use, perceived customizability, fitness for translation evaluation, and will-to-use for evaluation in the future.

[3]Results and exports are available only via explicit action to avoid mid-campaign decisions and bias.

| | Time | Ease | Cust. | Fit | Use |
|---|---|---|---|---|---|
| Pearmut | 11m | 9.0 | 7.4 | 9.2 | 8.4 |
| Appraise | 25m [2] | 4.4 | 6.2 | 7.6 | 4.8 |
| Potato | 23m [2] | 3.8 | 7.2 | 3.0 | 2.4 |
| Label Studio | 19m | 8.0 | 7.8 | 8.0 | 7.6 |
| Factgenie | 20m [1] | 5.4 | 7.0 | 6.4 | 5.6 |

Table 1: Average time (minutes) to setup minimal working annotation example and averaged answers on 11-point scale (0=worst, 5=middle, 10=best) across five researchers. The ● shows the number of failures to set up. See per-researcher breakdown in Table 4.

**Results.** While the sample size prevents statistical generalization, Table 1 shows a strong directional signal regarding the suitability of Pearmut for translation evaluation. Beyond speed and number of researchers giving up, Pearmut sacrifices generic flexibility (lower customizability) to provide a hyper-optimized experience for translation and other multilingual tasks. Label Studio, despite being close to Pearmut and being more flexible, does not have the appropriate mechanisms for managing a larger crowd of annotators, such as automated tutorials or assignment.

To test compatibility with modern engineering workflows, a similar task was assigned to an LLM-coding agent (see details in Appendix C) which succeeded in writing a setup script for Pearmut and Label Studio but not other tools.

## 5 Case Study with Annotators

We also conducted a study focused on the *annotators* rather than the *researchers*. Ten bilingual speakers annotated in total 540 translation outputs in English→{German, Czech×3, Hindi, Italian, Polish, Slovak, Spanish} in Pearmut and Appraise using ESA (Kocmi et al., 2024a). The annotation interface swapped each time after annotating three documents, with the starting interface randomized and the documents seen in the two interfaces different to avoid priming. On average, one session lasted an hour. See details in Appendix D.2.

**Results.** Annotators performed the task with Pearmut similarly fast and in-depth as in Appraise (as shown by number of annotated errors, separation between models, and inter-annotator agreement; results in Table 2, Appendix Table 6). When asked, the annotators attributed the much higher satisfaction to the interface being more responsive, the layout allowing for quick parallel reading, the option to switch between word- and character-level annotations, and to turn off approximate parallel character alignment, which some found distracting. These results corroborate Läubli et al. (2022) regarding the impact of mere interface on annotation performance.

| | Appraise | Pearmut |
|---|---|---|
| Time/item (s) | 144.86 ±23.89 | 124.38 ±19.30 |
| Time/char (ms) | 7.32 ±1.21 | 6.29 ±0.98 |
| Time/error (s) | 47.26 ±22.70 | 40.61 ±17.46 |
| Model A score | 91.94 ±3.02 | 95.42 ±2.01 |
| Model B score | 65.37 ±5.86 | 68.88 ±5.23 |
| Model C score | 33.56 ±4.92 | 33.43 ±5.03 |
| Model A errors/item | 0.8 ±0.32 | 0.5 ±0.21 |
| Model B errors/item | 3.8 ±0.91 | 3.4 ±0.89 |
| Model C errors/item | 7.1 ±1.43 | 6.5 ±1.08 |
| Speed (0 to 10) | 6.3 | 8.3 |
| Clarity (0 to 10) | 6.0 | 8.3 |
| Effort (0 to 10) | 6.0 | 7.8 |

Table 2: Annotation times, model results, and answers on an 11-point scale (0=worst, 10=best). Cells are averaged across ten annotators and their language pairs. The ± shows 95% t-distribution confidence intervals.

Lastly, Papi et al. (2025); Züfle et al. (2026) already used this tool with live annotators for diagnosing the errors in speech models and the WMT 2026 General Translation Shared Task (Kocmi et al., 2026) use Pearmut for large-scale human evaluation across many language pairs.

**Scalability.** We also measured the server response speed. The results in Appendix E show that Pearmut responds at least 3× faster to requests for showing the next evaluation item, showing management dashboard, and downloading results. The expected maximum number of concurrent users on cheap hardware at the same time is 12× higher than the 158 used by large-scale WMT 2025 evaluation spread out across two months (Kocmi et al., 2025), see Appendix E for computations.

## 6 Conclusion

We presented Pearmut, an evaluation platform that fulfills the needs of practitioners human-evaluating multilingual tasks and translation. The focus of this tool is to remove entry barriers and make it as easy to use as automatic metrics (e.g. SacreBLEU). It enables best practices (unified evaluation, statistical testing) rather than making practitioners introduce new annotation protocols. Through two case studies on practitioners and annotators we show the ease of use of this platform.

## Limitations

The number of participants in the two users studies appears modest (5+10) but it is much higher than other frameworks (0+6 for Federmann, 2010, 0+1 for Pei et al., 2022, 0+0 for Kasner et al., 2024).

While the goal of Pearmut is to provide accessible human evaluation platform, it does still require engineering skills, which might exclude some groups. Specifically, the practitioner needs to prepare the data in adequate format (campaign JSON file), which is then uploaded to the running instance of Pearmut. Setting up the instance might also require engineering skills, such as basic CLI commands or port forwarding.

## Ethics Statement

Expert participants gave an informed consent to participate in the two user studies with full experiment setup fully disclosed (i.e. no deception). The texts in the user studies were sourced from an existing human-evaluated dataset (WMT25, Kocmi et al., 2025) and thus deemed benign and exempt from ethics board approval.

The collected data does not contain annotator-identifiable information and user IDs are randomly generated strings, such as `calm-ligand-106`. This data is stored on the server and accessible only via a privileged link. It is up to the experiment managers to ensure that all annotators give proper consent.

Publicly available LLMs were used for post-editing the writing of this paper and for coding assistance.

## Acknowledgments

Vilém Zouhar gratefully acknowledges the support of the Google PhD Fellowship.

Thanks to Niyati Bafna and Ahrii Kim for invaluable feedback during the development of this tool and paper. Thanks to Ahrii Kim, Sankalan Pal Chowdhury, KV Aditya Srivatsa, Kateřina Stodolová, Tommaso Felice Banfi, Ona de Gibert, Ivan Kartáč, Kristýna Onderková, Maike Züfle, and the annonymous participants for their contributions in the user studies.

## References

Abubakar Abid, Ali Abdalla, Ali Abid, Dawood Khan, Abdulrahman Alfozan, and James Zou. 2019. Gradio: Hassle-free sharing and testing of ml models in the wild.

Farhad Akhbardeh, Arkady Arkhangorodsky, Magdalena Biesialska, Ondřej Bojar, Rajen Chatterjee, Vishrav Chaudhary, Marta R. Costa-jussa, Cristina España-Bonet, Angela Fan, Christian Federmann, Markus Freitag, Yvette Graham, Roman Grundkiewicz, Barry Haddow, Leonie Harter, Kenneth Heafield, Christopher Homan, Matthias Huck, Kwabena Amponsah-Kaakyire, Jungo Kasai, Daniel Khashabi, Kevin Knight, Tom Kocmi, Philipp Koehn, Nicholas Lourie, Christof Monz, Makoto Morishita, Masaaki Nagata, Ajay Nagesh, Toshiaki Nakazawa, Matteo Negri, Santanu Pal, Allahsera Auguste Tapo, Marco Turchi, Valentin Vydrin, and Marcos Zampieri. 2021. Findings of the 2021 Conference on Machine Translation (WMT21). In *Proceedings of the Sixth Conference on Machine Translation*, pages 1–88, Online.

James E Baker. 1985. Adaptive selection methods for genetic algorithms. In *Proceedings of the 1st International Conference on Genetic Algorithms*, pages 101–111, Pittsburgh, PA, USA.

Ondřej Bojar, Rajen Chatterjee, Christian Federmann, Yvette Graham, Barry Haddow, Matthias Huck, Antonio Jimeno Yepes, Philipp Koehn, Varvara Logacheva, Christof Monz, Matteo Negri, Aurélie Névéol, Mariana Neves, Martin Popel, Matt Post, Raphael Rubino, Carolina Scarton, Lucia Specia, Marco Turchi, Karin Verspoor, and Marcos Zampieri. 2016. Findings of the 2016 Conference on Machine Translation. In *Proceedings of the First Conference on Machine Translation: Volume 2, Shared Task Papers*, pages 131–198, Berlin, Germany.

Carlo Bonferroni. 1936. Teoria statistica delle classi e calcolo delle probabilita. *Pubblicazioni del R Istituto superiore di scienze economiche e commericiali di Firenze* 8:3–62.

Wei-Lin Chiang, Lianmin Zheng, Ying Sheng, Anastasios Nikolas Angelopoulos, Tianle Li, Dacheng Li, Hao Zhang, Banghua Zhu, Michael Jordan, Joseph E. Gonzalez, and Ion Stoica. 2024. Chatbot Arena: An Open Platform for Evaluating LLMs by Human Preference.

Christian Federmann. 2010. Appraise: An Open-Source Toolkit for Manual Phrase-Based Evaluation of Translations.

Christian Federmann. 2018. Appraise Evaluation Framework for Machine Translation. In *Proceedings of the 27th International Conference on Computational Linguistics: System Demonstrations*, pages 86–88, Santa Fe, New Mexico.

Markus Freitag, George Foster, David Grangier, Viresh Ratnakar, Qijun Tan, and Wolfgang Macherey. 2021. Experts, Errors, and Context: A Large-Scale Study of Human Evaluation for Machine Translation.

*Transactions of the Association for Computational Linguistics* 9:1460–1474.

Yvette Graham, Timothy Baldwin, and Nitika Mathur. 2015. Accurate Evaluation of Segment-level Machine Translation Metrics. In *Proceedings of the 2015 Conference of the North American Chapter of the Association for Computational Linguistics: Human Language Technologies*, pages 1183–1191, Denver, Colorado.

Yvette Graham, Timothy Baldwin, Alistair Moffat, and Justin Zobel. 2013a. Continuous Measurement Scales in Human Evaluation of Machine Translation. In *Proceedings of the 7th Linguistic Annotation Workshop and Interoperability with Discourse*, pages 33–41, Sofia, Bulgaria.

Yvette Graham, Timothy Baldwin, Alistair Moffat, and Justin Zobel. 2013b. Crowd-Sourcing of Human Judgments of Machine Translation Fluency. In *Proceedings of the Australasian Language Technology Association Workshop 2013 (ALTA 2013)*, pages 16–24, Brisbane, Australia.

Yvette Graham, Timothy Baldwin, Alistair Moffat, and Justin Zobel. 2017. Can machine translation systems be evaluated by the crowd alone. *Natural Language Engineering* 23(1):3–30.

Zdeněk Kasner, Ondrej Platek, Patricia Schmidtova, Simone Balloccu, and Ondrej Dusek. 2024. factgenie: A Framework for Span-based Evaluation of Generated Texts. In *Proceedings of the 17th International Natural Language Generation Conference: System Demonstrations*, pages 13–15, Tokyo, Japan.

Daniel Khashabi, Gabriel Stanovsky, Jonathan Bragg, Nicholas Lourie, Jungo Kasai, Yejin Choi, Noah A. Smith, and Daniel Weld. 2022. GENIE: Toward Reproducible and Standardized Human Evaluation for Text Generation. In *Proceedings of the 2022 Conference on Empirical Methods in Natural Language Processing*, pages 11444–11458, Abu Dhabi, United Arab Emirates.

Douwe Kiela, Max Bartolo, Yixin Nie, Divyansh Kaushik, Atticus Geiger, Zhengxuan Wu, Bertie Vidgen, Grusha Prasad, Amanpreet Singh, Pratik Ringshia, Zhiyi Ma, Tristan Thrush, Sebastian Riedel, Zeerak Waseem, Pontus Stenetorp, Robin Jia, Mohit Bansal, Christopher Potts, and Adina Williams. 2021. Dynabench: Rethinking Benchmarking in NLP. In *Proceedings of the 2021 Conference of the North American Chapter of the Association for Computational Linguistics: Human Language Technologies*, pages 4110–4124, Online.

Leonard Kleinrock. 1975. *Queueing Systems, Volume 1: Theory*. Wiley-Interscience.

Tom Kocmi, Ekaterina Artemova, Eleftherios Avramidis, Rachel Bawden, Ondřej Bojar, Konstantin Dranch, Anton Dvorkovich, Sergey Dukanov, Mark Fishel, Markus Freitag, Thamme Gowda, Roman Grundkiewicz, Barry Haddow, Marzena Karpinska, Philipp Koehn, Howard Lakougna, Jessica Lundin, Christof Monz, Kenton Murray, Masaaki Nagata, Stefano Perrella, Lorenzo Proietti, Martin Popel, Maja Popović, Parker Riley, Mariya Shmatova, Steinthór Steingrímsson, Lisa Yankovskaya, and Vilém Zouhar. 2025. Findings of the WMT25 General Machine Translation Shared Task: Time to Stop Evaluating on Easy Test Sets. In *Proceedings of the Tenth Conference on Machine Translation*, pages 355–413, Suzhou, China.

Tom Kocmi, Ekaterina Artemova, Eleftherios Avramidis, Rachel Bawden, Ondřej Bojar, Sergey Dukanov, Anton Dvorkovich, Mark Fishel, Markus Freitag, Samuel Frontull, Thamme Gowda, Roman Grundkiewicz, Barry Haddow, Stan Kharevich, Philipp Koehn, Zheng Li, Jean Maillard, Christof Monz, Alexander Murauski, Kenton Murray, Masaaki Nagata, Stefano Perrella, Martin Popel, Maja Popović, Lorenzo Proietti, Sara Rajaee, Parker Riley, Mariya Shmatova, Steinþór Steingrímsson, Lisa Yankovskaya, and Vilém Zouhar. 2026. Findings of the WMT26 General Machine Translation Shared Task: Contrastive Dynamic Human Evaluation at Scale. In *To appear in Proceedings of the Eleventh Conference on Machine Translation*.

Tom Kocmi, Vilém Zouhar, Eleftherios Avramidis, Roman Grundkiewicz, Marzena Karpinska, Maja Popović, Mrinmaya Sachan, and Mariya Shmatova. 2024a. Error Span Annotation: A Balanced Approach for Human Evaluation of Machine Translation. In *Proceedings of the Ninth Conference on Machine Translation*, pages 1440–1453, Miami, Florida, USA.

Tom Kocmi, Vilém Zouhar, Christian Federmann, and Matt Post. 2024b. Navigating the Metrics Maze: Reconciling Score Magnitudes and Accuracies. In *Proceedings of the 62nd Annual Meeting of the Association for Computational Linguistics (Volume 1: Long Papers)*, pages 1999–2014, Bangkok, Thailand.

Alon Lavie, Greg Hanneman, Sweta Agrawal, Diptesh Kanojia, Chi-Kiu Lo, Vilém Zouhar, Frederic Blain, Chrysoula Zerva, Eleftherios Avramidis, Sourabh Deoghare, Archchana Sindhujan, Jiayi Wang, David Ifeoluwa Adelani, Brian Thompson, Tom Kocmi, Markus Freitag, and Daniel Deutsch. 2025. Findings of the WMT25 Shared Task on Automated Translation Evaluation Systems: Linguistic Diversity is Challenging and References Still Help. In *Proceedings of the Tenth Conference on Machine Translation*, pages 436–483, Suzhou, China.

Samuel Läubli, Patrick Simianer, Joern Wuebker, Geza Kovacs, Rico Sennrich, and Spence Green. 2022. The impact of text presentation on translator perfor-mance. *Target. International Journal of Translation Studies* 34(2):309-342.

Benjamin Marie, Atsushi Fujita, and Raphael Rubino. 2021. Scientific Credibility of Machine Translation Research: A Meta-Evaluation of 769 Papers. In *Proceedings of the 59th Annual Meeting of the Association for Computational Linguistics and the 11th International Joint Conference on Natural Language Processing (Volume 1: Long Papers)*, pages 7297–7306, Online.

Nitika Mathur, Timothy Baldwin, and Trevor Cohn. 2017. Sequence Effects in Crowdsourced Annotations. In *Proceedings of the 2017 Conference on Empirical Methods in Natural Language Processing*, pages 2860–2865, Copenhagen, Denmark.

Jekaterina Novikova, Ondřej Dušek, and Verena Rieser. 2018. RankME: Reliable Human Ratings for Natural Language Generation. In *Proceedings of the 2018 Conference of the North American Chapter of the Association for Computational Linguistics: Human Language Technologies, Volume 2 (Short Papers)*, pages 72–78, New Orleans, Louisiana.

Sara Papi, Javier Garcia Gilabert, Zachary Hopton, Vilém Zouhar, Carlos Escolano, Gerard I. Gállego, Jorge Iranzo-Sánchez, Ahrii Kim, Dominik Macháček, Patricia Schmidtova, and Maike Züfle. 2025. Hearing to Translate: The Effectiveness of Speech Modality Integration into LLMs. arXiv: 2512.16378 [cs.CL].

Jiaxin Pei, Aparna Ananthasubramaniam, Xingyao Wang, Naitian Zhou, Apostolos Dedeloudis, Jackson Sargent, and David Jurgens. 2022. POTATO: The Portable Text Annotation Tool. In *Proceedings of the 2022 Conference on Empirical Methods in Natural Language Processing: System Demonstrations*, pages 327–337, Abu Dhabi, UAE.

Matt Post. 2018. A Call for Clarity in Reporting BLEU Scores. In *Proceedings of the Third Conference on Machine Translation: Research Papers*, pages 186–191, Brussels, Belgium.

Keisuke Sakaguchi, Matt Post, and Benjamin Van Durme. 2014. Efficient Elicitation of Annotations for Human Evaluation of Machine Translation. In *Proceedings of the Ninth Workshop on Statistical Machine Translation*, pages 1–11, Baltimore, Maryland, USA.

Keisuke Sakaguchi and Benjamin Van Durme. 2018. Efficient Online Scalar Annotation with Bounded Support. In *Proceedings of the 56th Annual Meeting of the Association for Computational Linguistics (Volume 1: Long Papers)*, pages 208–218, Melbourne, Australia.

Belén Saldías Fuentes, George Foster, Markus Freitag, and Qijun Tan. 2022. Toward More Effective Human Evaluation for Machine Translation. In *Proceedings of the 2nd Workshop on Human Evaluation of NLP Systems (HumEval)*, pages 76–89, Dublin, Ireland.

Yixiao Song, Parker Riley, Daniel Deutsch, and Markus Freitag. 2025. Enhancing human evaluation in machine translation with comparative judgment.

Maxim Tkachenko, Mikhail Malyuk, Andrey Holmanyuk, and Nikolai Liubimov. no date. Label Studio: Data labeling software. Open source software available from https://github.com/HumanSignal/label-studio.

Vilém Zouhar, Peng Cui, and Mrinmaya Sachan. 2025a. How to Select Datapoints for Efficient Human Evaluation of NLG Models? *Transactions of the Association for Computational Linguistics* 13:1789-1811.

Vilém Zouhar, Tom Kocmi, and Mrinmaya Sachan. 2025b. AI-Assisted Human Evaluation of Machine Translation. In *Proceedings of the 2025 Conference of the Nations of the Americas Chapter of the Association for Computational Linguistics: Human Language Technologies (Volume 1: Long Papers)*, pages 4936–4950, Albuquerque, New Mexico.

Maike Züfle, Ondrej Klejch, Nicholas Sanders, Jan Niehues, Alexandra Birch, and Tsz Kin Lam. 2026. F-Actor: Controllable Conversational Behaviour in Full-Duplex Models. arXiv: 2601.11329 [cs.CL].

| | Pearmut | Factgenie | Appraise | Potato | Label Studio | ChatbotArena | Spreadsheet |
|---|---|---|---|---|---|---|---|
| Focus | NLG/MT | NLG | MT | General | General | LLM | General |
| Protocols | DA, ESA, MQM, contrastive | Spans, Ranking | DA, ESA, MQM | Classif., Spans | Classif., Spans, Multi-modal | Pairwise comparison | Limited |
| Setup Complexity | Low | Low | High | Low | Medium | High | Low on small scale |
| Quality Control | Yes | Limited | Limited | Limited | No | Yes | No |
| Assignment | Static, Automatic, Dynamic | Static | Static | Static, Automatic | Manual | Dynamic | Static |

Table 3: Comparison of Pearmut against established annotation and evaluation interfaces.

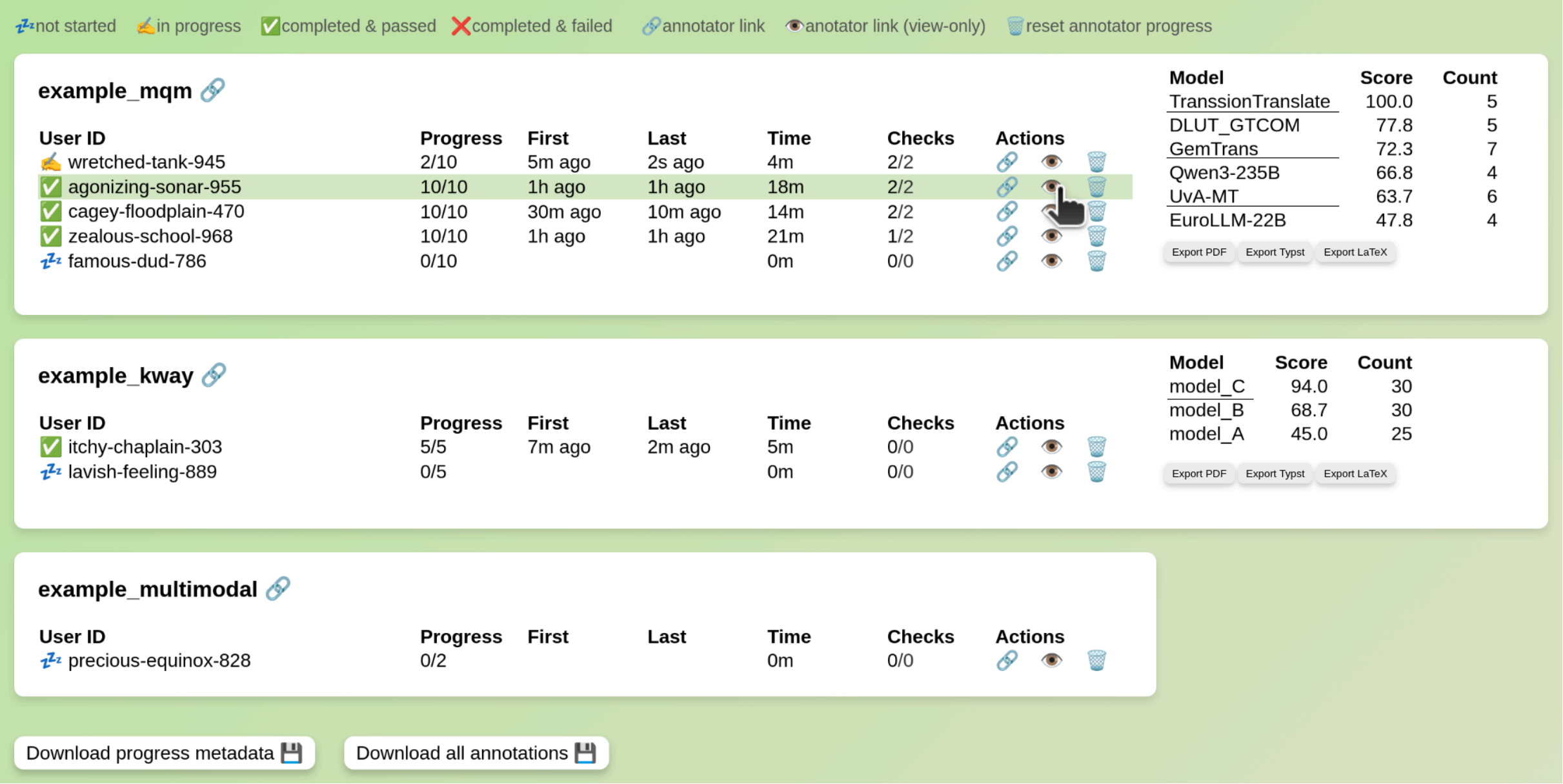

Figure 2: Screenshot of the Pearmut dashboard interface with mock model results. Horizontal lines show statistically significant model averages (two-sided t-test with related samples with p-value under 5%).

# A Technical Details

This appendix describes various technical and implementation aspects of Pearmut. For example, one aspect in Pearmut design was accessibility. A simulated example of color-blind perception of Pearmut is shown in Figure 4.

## A.1 Architecture

The tool consists of three components which are accessible after running Pearmut:
1. server serving static files and API requests,
2. frontend annotation templates,
3. frontend dashboard for monitoring.

In contrast to other platforms, Pearmut does not use server-rendered templates. Instead, it serves a static page with an accompanying user-side code that queries the server for the data. This leads to higher responsiveness and allows for more flexibility when implementing new protocols.

The server-side is built with FastAPI in Python and the front-end with TypeScript with jQuery built with Webpack, though the frontend is easily exchangeable. The pip package contains the frontend pre-built. The runtime data is stored in-memory for low-latency read access, backed by an append-only log file on disk. Every write operation is synchronously flushed to the log, ensuring data durability in the event of a process crash. This architecture avoids the overhead of a heavy database management system while maintaining reliability for concurrent scales. The campaign definitions and annotation outputs are JSON files for easy analysis processing.

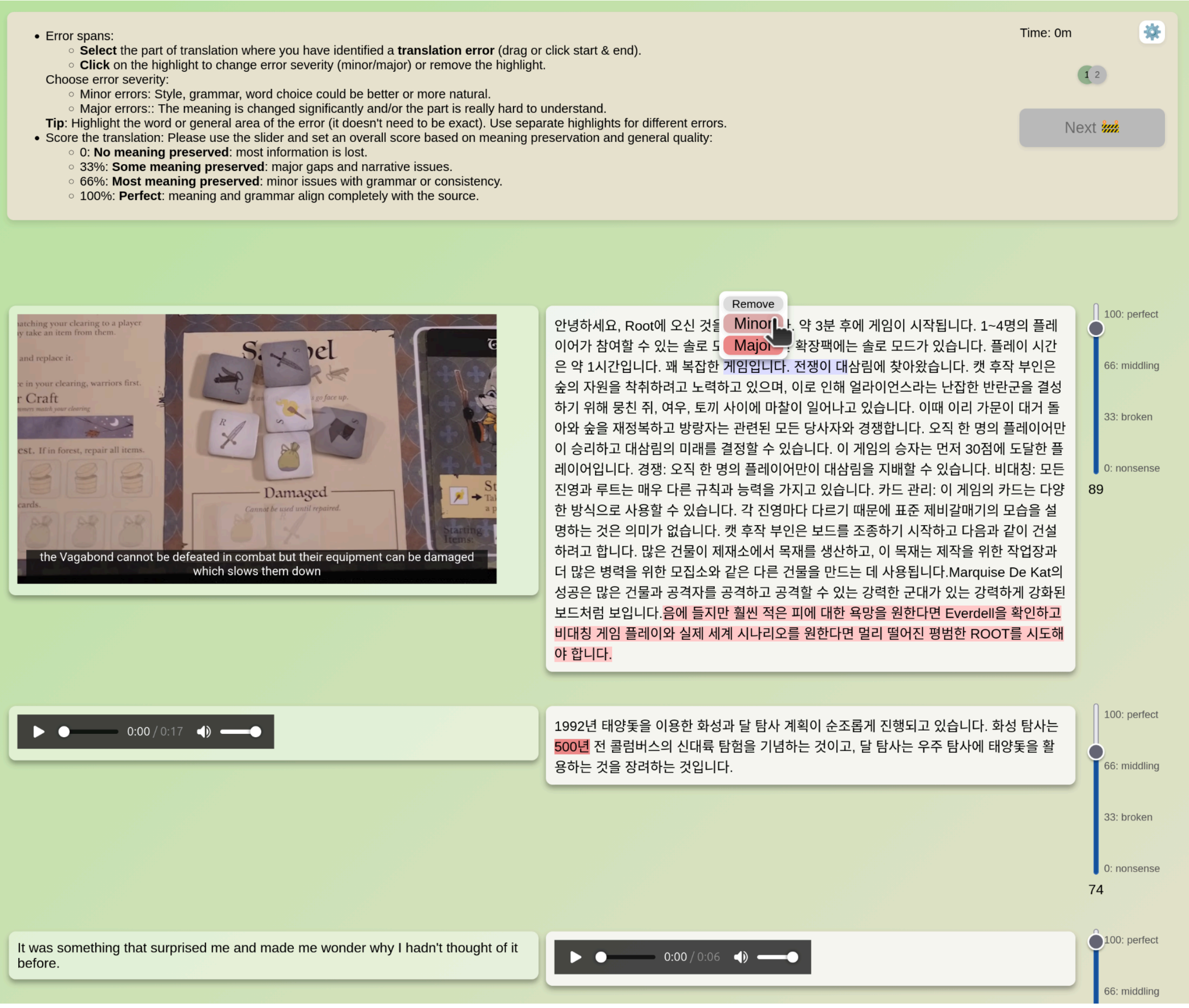

Figure 3: Screenshot of the Pearmut annotation interface with multimodal inputs or outputs with Hangul script (Korean). From top: speech translation based on a video, speech translation based on audio, speech synthesis/output based on text. References are supported but not shown for brevity.

**Security Note.** The magic links (one-step passwordless authentication) contain 96-bit random tokens. This approach prioritizes low friction for crowd-workers over strict access control. The platform is primarily intended for non-sensitive public datasets; for proprietary or sensitive data, we recommend deploying Pearmut on an internal network or behind an authentication proxy (e.g., VPN).

## A.2 Dynamic Assignment

One option is `dynamic_models`, which samples multiple models according to the current ranking $\frac{1}{\text{rank}}$. This happens after a cold start phase, `dynamic_coldstart`.

For contrastive evaluation (comparing multiple model outputs at the same time, Figure 1), Pearmut supports comparing systems that are close in performance. This ensures that the likely-to-be state-of-the-art models are compared against each other and maximizes signal per annotation, rather than wasting budget comparing them with a distinctively weak baseline.

The various assignment modes are shown in Figure 5. Consider an example with `dynamic_coldstart=5`, and models A, B, C, which are sorted in descending quality. We first evaluate all models on 5 evaluation items. Then we find out that their averages so far are 90, 49, and 50 respectively. In this next stage, we will therefore prioritize evaluating models A and C. At first glance this seems incorrect as B > C. However, we will soon start focusing on evaluating models A and B. This is because by evaluating C, its average will drop below the average of B, or that evaluating B will raise its average above that of C.

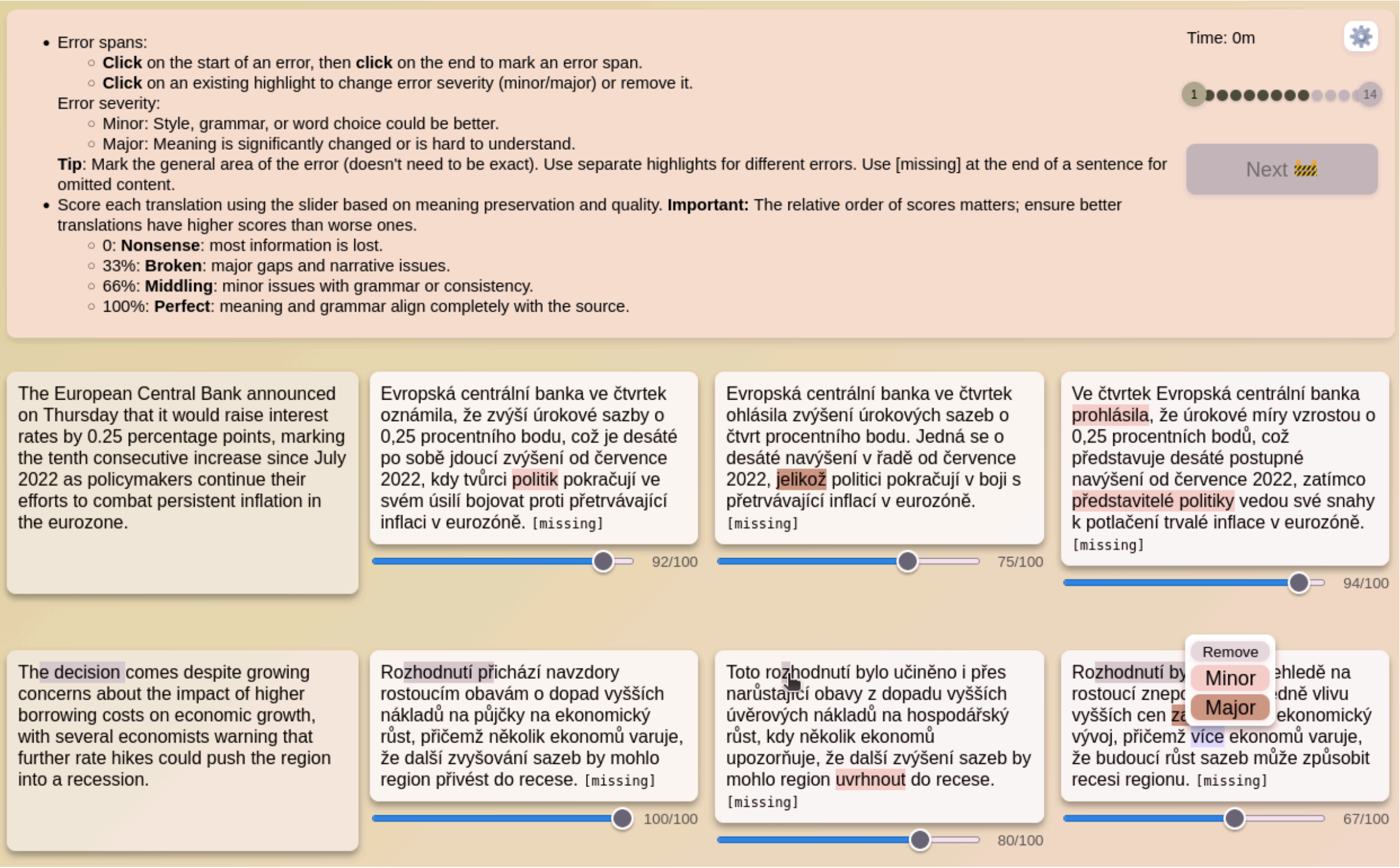

Figure 4: Deuteranopia-colorblind simulated screenshot (Figure 1) of the Pearmut annotation interface contrastive ESA together with annotation guidelines.

This approach can be conceptualized as a best arm identification in a multi-armed bandit problem. The models are arms, and pulling an arm means evaluating a single item with this model. Because pulling an arm is expensive, we wish to find the best arm (model) with as few arm pulls as possible. By using `dynamic_coldstart` we first obtain reasonable estimates on the arm averages and by sampling $\frac{1}{\text{rank}}$ we perform exploration-focused multi-armed bandit optimization (Baker, 1985). The proper mathematical and empirical analysis of this approach is beyond the scope of this paper, which describes the platform on which these assignments can be built. However, we note that this approach is *consistent*, i.e. with enough evaluation samples, it will converge to the true model ranking because there exists $\delta > 0$ such that $\frac{1}{\text{rank}} > \delta > 0$.

While dynamic assignment improves data efficiency for identifying the top system, it introduces selection bias (better systems are sampled more). Consequently, standard frequentist significance tests (like the t-tests in the dashboard) are strictly valid only for the `single-stream` or `task-based` (randomized) assignments. Practitioners using dynamic assignment should export the data and apply corrections (e.g., inverse propensity weighting or Bonferroni correction, Bonferroni, 1936) for reporting.

### A.3 Dashboard and Statistical Testing

A sample dashboard is shown in Figure 2. Intentionally, showing and exporting the model results is hidden behind a button that needs to be explicitly clicked. This way, we wish to prevent accidental mid-campaign decisions that could bias the results.

The aggregate results (model ranking in Figure 2) contains occasional horizontal lines. These mark that the models are statistically different based on a two-sided t-test on related samples (annotated for both models) with 5% threshold. We chose this parametric statistical test as a reasonable default as its requirements are that the item-wise differences between the models are independent and normally distributed. This improves standards in scientific reporting because of the pre-defined default serving as a partial pre-registration. Not using this default is easily possible but should always require a sound justification.

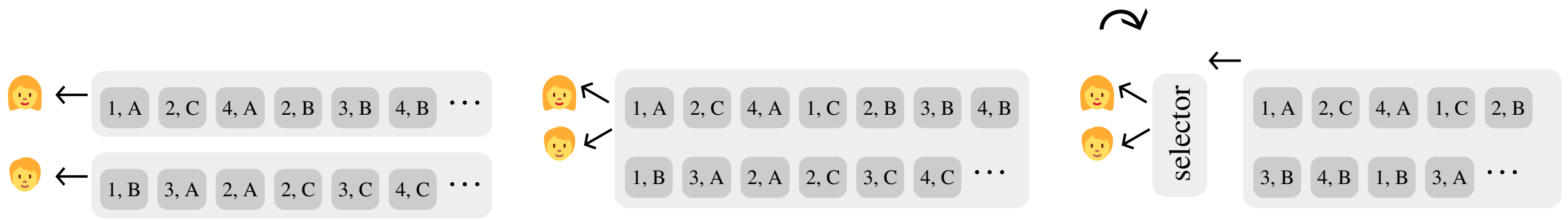

(a) Task-based assignment (b) Single-stream assignment (c) Dynamic assignment

Figure 5: Illustration of various methods of assigning evaluation items to annotators. The single-stream assignment (c) randomly samples from a set of items. The dynamic assignment (c) uses the already collected annotations to select the next evaluation item to distribute.

### A.4 Tutorials and Attention Checks

Pearmut aids in ensuring the quality of annotations by implementing tutorial evaluation items and attention checks. The researcher can specify that a particular translation has a specific score, a specific error span, or has a higher score than an intentionally bad translation. A tutorial item is simply an attention check which does not let the annotator pass until conditions are met and when they are not met, a warning message is shown. Attention checks, on the other hand, simply register the pass or failure in the dashboard but do not notify the annotator. Based on successful passing of enough attention checks, the annotator is shown an *accept* or *reject* token, which can integrate with crowdsourcing services.

## B Case Study with Researchers

The participants were shown the following instructions:

*Your task is to human-evaluate the quality of the following translations:*

```
Source:  The quick brown fox jumped over the lazy dog.
Model-A: Der schnell braun Fuchs springte über das faul Hund.
Model-B: Der schnelle braune Fuchs sprang über den faulen Hund.

Source:  The European Central Bank announced on Thursday that it would raise interest
rates by 0.25 percentage points, marking the tenth consecutive increase since July 2022
as policymakers continue their efforts to combat persistent inflation in the eurozone.
Model-A: Die Europäisch Zentral Bank ankündigte am Donnerstag, dass es würde heben
Interesse Raten bei 0.25 Prozent Punkten, markierend die zehnte konsekutive Erhöhung seit
Juli 2022 als Politikmacher weitermachen ihre Anstrengungen zu kämpfen persistente
Inflation in die Eurozone.
Model-B: Die Europäische Zentralbank gab am Donnerstag bekannt, dass sie die Leitzinsen
um 0,25 Prozentpunkte anheben werde. Dies markiert die zehnte Erhöhung in Folge seit Juli
2022, da die Entscheidungsträger ihre Bemühungen zur Bekämpfung der hartnäckigen
Inflation in der Eurozone fortsetzen
```

*Use the following tools to set up an annotation campaign to find out if Model-A or Model-B is better.*

- *Appraise github.com/AppraiseDev/Appraise*
- *Potato github.com/davidjurgens/potato*
- *Pearmut github.com/zouharvi/pearmut*
- *Label Studio github.com/HumanSignal/label-studio*
- *Factgenie github.com/ufal/Factgenie*

*For each, set up the evaluation campaign, send "instructions" to your annotators who will annotate the results, and then interpret the results to conclude your mock "study". Time checkpoints will be measured for (1) installing the software, (2) setting up the campaign, (3) annotations complete, and (4) obtaining results.*

*After finishing please evaluate each on scale of 0 (worst) to 10 (best):*

1. *How easy was the tool to use?*
2. *How customizable is the tool?*
3. *How fitting is the tool for translation evaluation?*
4. *How likely would you use the tool for your next study of translation evaluation?*

All participants in this study were NLP researchers but without particular experience with any of the annotation platforms. The order of the platforms to setup was randomized. If a particular step was taking longer than 10 minutes, they could ask for guidance with the particular step. The results per researcher are shown in Table 4.

| | Tool | Install | Data | Annot. | Collect | Total | Ease | Cust. | Fit | Use |
|---|---|---|---|---|---|---|---|---|---|---|
| | Pearmut | 1m | 17m | 1m | 1m | 21m | 9 | 9 | 10 | 10 |
| | Appraise | 12m | 30m ✕ | ✕ | ✕ | ✕ | 5 | 9 | 10 | 8 |
| **R**A | Potato | 3m | 23m ✕ | ✕ | ✕ | ✕ | 0 | 9 | 0 | 0 |
| | Label Studio | 4m | 12m | 1m | 1m | 18m | 10 | 10 | 10 | 10 |
| | Factgenie | 35m | 34m ✕ | ✕ | ✕ | ✕ | 6 | 10 | 10 | 5 |
| | Pearmut | 2m | 5m | 1m | 1m | 9m | 9 | 7 | 9 | 9 |
| | Appraise | 5m | 10m | 2m | 2m | 19m | 6 | 6 | 7 | 6 |
| **R**B | Potato | 2m | 15m ⚠ | 1m | 5m | 23m | 4 | 8 | 5 | 4 |
| | Label Studio | 5m | 8m | 2m | 2m | 17m | 8 | 9 | 8 | 9 |
| | Factgenie | 2m | 20m | 1m | 2m | 25m | 5 | 6 | 5 | 5 |
| | Pearmut | 1m | 3m | 1m | 1m | 7m | 10 | 8 | 10 | 10 |
| | Appraise | 3m | 6m | 2m | 2m | 13m | 7 | 5 | 8 | 7 |
| **R**C | Potato | 2m | 6m ✕ | ✕ | ✕ | ✕ | 2 | 5 | 2 | 0 |
| | Label Studio | 3m | 7m | 2m | 2m | 14m | 9 | 9 | 9 | 9 |
| | Factgenie | 2m | 12m ⚠ | 2m | 2m | 18m | 8 | 9 | 8 | 8 |
| | Pearmut | 4m | 5m | 1m | 1m | 10m | 8 | 7 | 7 | 7 |
| | Appraise | 18m | 19m ❓ | 2m | 6m | 45m | 2 | 3 | 3 | 3 |
| **R**D | Potato | 6m ❓ | 14m | 1m | 1m | 22m | 6 | 6 | 1 | 3 |
| | Label Studio | 21m ❓ | 9m | 1m | 2m | 33m | 6 | 7 | 7 | 6 |
| | Factgenie | 4m | 15m ❓ ⚠ | 1m | 2m | 22m | 2 | 6 | 5 | 6 |
| | Pearmut | 1m | 9m | 1m | 1m | 11m | 9 | 6 | 10 | 6 |
| | Appraise | 2m | 24m ✕ | ✕ | ✕ | ✕ | 2 | 8 | 10 | 0 |
| **R**E | Potato | 1m | 22m | 1m | 1m | 25m | 7 | 8 | 7 | 5 |
| | Label Studio | 4m | 12m | 1m | 1m ❓ | 17m | 7 | 4 | 6 | 4 |
| | Factgenie | 2m | 14m | 1m | 1m | 17m | 6 | 4 | 4 | 4 |

Table 4: Per-researcher (anonymized to a single letter) time durations per checkpoint and answer on 11-point scale (0=worst, 5=middle, 10=best). The ⚠ marks a problem, ❓ that help was provided due to time constraints, and ✕ that a problem occurred that prevented continuing and the researcher gave up.

## C Evaluation of Agentic Workflow Compatibility

The following instructions were given to Gemini 3 pro with thinking model, with the tool name and documentation URL changed:

*Your task is to prepare a human evaluation campaign to assess the quality of translations from English to German:*

```
Source:  The quick brown fox jumped over the lazy dog.
Model-A: Der schnell braun Fuchs springte über das faul Hund.
Model-B: Der schnelle braune Fuchs sprang über den faulen Hund.

Source:  The European Central Bank announced on Thursday that it would raise interest
rates by 0.25 percentage points, marking the tenth consecutive increase since July 2022
as policymakers continue their efforts to combat persistent inflation in the eurozone.
```

```
Model-A: Die Europäisch Zentral Bank ankündigte am Donnerstag, dass es würde heben
Interesse Raten bei 0.25 Prozent Punkten, markierend die zehnte konsekutive Erhöhung seit
Juli 2022 als Politikmacher weitermachen ihre Anstrengungen zu kämpfen persistente
Inflation in die Eurozone.
Model-B: Die Europäische Zentralbank gab am Donnerstag bekannt, dass sie die Leitzinsen
um 0,25 Prozentpunkte anheben werde. Dies markiert die zehnte Erhöhung in Folge seit Juli
2022, da die Entscheidungsträger ihre Bemühungen zur Bekämpfung der hartnäckigen
Inflation in der Eurozone fortsetzen
```

*Write a bash script that installs the tool Pearmut/Appraise/Potato/Label Studio/Factgenie and generates a link that we can give to annotators to annotate. Read the documentation for the tool first: https://…*
*Then write another bash script that retrieves the human-annotated data.*

The experiment served as a stress test for API predictability and documentation readiness. The agent successfully synthesized valid configuration scripts for Pearmut and Label Studio. In contrast, for tools like Appraise and Factgenie, the model hallucinated non-existent commands or attempted incorrect import strategies (e.g., treating a standalone application as a Python library).

These failures indicate that those tools rely on complex, non-standard architectural patterns or knowledge not explicitly captured in the documentation. Pearmut's success validates that its CLI design aligns with standard package management practices (pip-installable, JSON configuration), ensuring it can be easily integrated into modern, AI-assisted research pipelines without manual debugging.

## D Case Study with Annotators

### D.1 Instructions

The annotation interface was set up by the authors and the annotators were given two links: one for Pearmut and one for Appraise. The annotation guidelines were:

*Your task is to annotate several translations in two interfaces: Pearmut and Appraise using the Error Span Annotation protocol. You will be shown a sequence of documents in which you will have to mark errors.*
*Instructions also shown during the annotation:*

- *Error spans:*
  - *Click on the start of an error, then click on the end to mark an error span.*
  - *Click/hover on an existing highlight to change error severity (minor/major) or remove it.*
- *Error severity:*
  - *Minor: Style, grammar, or word choice could be better.*
  - *Major: Meaning is significantly changed or is hard to understand.*
- *Tip: Mark the general area of the error (doesn't need to be exact). Use separate highlights for different errors. Use* `[missing]` *at the end of a sentence for omitted content.*
- *Score each translation using the slider based on meaning preservation and quality. Important: The relative order of scores matters; ensure better translations have higher scores than worse ones.*
  - *0: Nonsense: most information is lost.*
  - *33%: Broken: major gaps and narrative issues.*
  - *66%: Middling: minor issues with grammar or consistency.*
  - *100%: Perfect: meaning and grammar align completely with the source.*

*After three documents (screens, each document has three texts) in one interface, you will transition to another interface and then back again until you complete nine documents in each interface.*
*At the end, please evaluate the two tools each on scale of 0 (worst) to 10 (best):*

- *How fast was the tool to use?*
- *How clear was the tool to use?*
- *How much effort, excluding the evaluation task itself, was required to interact with the interface?*

### D.2 Setup Details

None of the annotators worked inside of any of the two tools before and they were shown instructions from Appendix D.1. We took English source data (three documents for each tool, each with three segments) randomly sampled from WMT 2025 (Kocmi et al., 2025) and created three different translations using Gemini 3 pro, with increasing amount of errors (see Appendix D.2). This way, we know apriori what the final model ranking should be (A > B > C). The annotations were done with the Error Span Annotation (Kocmi et al., 2024a) protocol. This protocol was chosen because it is implemented in both tools and the goal of the experiment is to show that functionally, Pearmut can handle the same task as the current go-to tool in translation evaluation. The within-subject design was chosen intentionally so that the 11-point assessments are not biased based on the participant selection.

We chose documents with 3 segments per document and created 3 translations of varying quality. The annotator first annotated the 3 translations of one document in one interface, then switched to the other interface, and so on. The translations with intentionally varying quality were done with a prompt to Gemini 3 pro:

*Translate the following JSON data into Czech. For each source, output 3 translations: (A) a perfect translation, (B) a poor translation with minor mistakes, and (C) a translation with major mistakes. Output the data in the same JSON format, just adding the translations into the "tgt" key as:*

```
"tgt": {
    "A": "perfect translation",
    "B": "poor translation with slight mistakes",
    "C": "translation with noticeable mistakes"
}
```

See example translations created this way in Table 5. Lastly, in this annotation study, we measured the time externally as the internal time measurements might differ between Appraise and Pearmut.

### D.3 Results

We show the inter-annotator agreement for three English→Czech annotators in Table 6, compared between Appraise and Pearmut. We chose Kendall's $\tau_b$ for group-by-item following WMT metrics shared task (Lavie et al., 2025) because in this grouping, we care about ranking. However, for global and group-by-model we care about the scale, which is why we chose Pearson. The correlations vary greatly, which is due to the modest scale of the study (models, items, annotators). Therefore, we only draw the conclusion that the correlations, some signal of annotation quality, are comparable.

In Figure 6 we show a time diagram of annotator actions, which is only possible with detailed logging that is often missing in similar tools. This reveals in which points annotators decide on annotating spans and scoring three segments within a document.

| English source | Slovak | German | Hindi |
|---|---|---|---|
| Bedtime story time! What's up with T9 keyboards? Why do 7 and 9 have 4 letters, but others 3. Why those two? Why not assign 1 some letters? Is 0 always been space? | **A:** Čas na rozprávku na dobrú noc! Čo sa deje s klávesnicami T9? Prečo majú 7 a 9 štyri písmená, ale ostatné len tri? Prečo práve tieto dve? Prečo nepriradiť nejaké písmená jednotke? Bola 0 vždy medzera? | **A:** Zeit für eine Gutenachtgeschichte! Was hat es eigentlich mit T9-Tastaturen auf sich? Warum haben die 7 und die 9 vier Buchstaben, die anderen aber nur drei? Warum gerade diese beiden? Warum weist man der 1 keine Buchstaben zu? War die 0 schon immer das Leerzeichen? | **A:** सोने के समय की कहानी! T9 कीबोर्ड्स का क्या चक्कर है? 7 और 9 में 4 अक्षर क्यों होते हैं, जबकि बाकियों में 3? वही दोनों क्यों? 1 को कुछ अक्षर क्यों नहीं दिए गए? क्या 0 हमेशा से स्पेस के लिए था? |
| | **B:** Čas na posteľný príbeh! Čo je s klávesnicami T9? Prečo 7 a 9 majú 4 písmená, ale iné 3. Prečo tie dve? Prečo nepriradiť 1 nejaké písmená? Je 0 vždy medzerou? | **B:** Bettzeit Geschichtszeit! Was ist los mit T9 Keyboards? Warum haben 7 und 9 vier Buchstaben, aber andere 3. Wieso diese zwei? Warum nicht der 1 manche Buchstaben zuweisen? Ist 0 immer Platz gewesen? | **B:** बिस्तर के समय की कहानी! T9 कीबोर्ड के साथ क्या हो रहा है? 7 और 9 में 4 अक्षर क्यों हैं, लेकिन दूसरों में 3. वो दो क्यों? 1 को कुछ पत्र क्यों नहीं सौंपते? क्या 0 हमेशा जगह रहा है? |
| | **C:** Čas na príbeh spánku! Čo je hore s T9 klávesmi? Prečo robia 7 a 9 mať 4 listy, ale iné 3. Prečo tamtie dva? Prečo nie priradiť 1 nejaké listy? Je 0 vždy bola vesmír? | **C:** Schlafenszeit Geschichte Zeit! Was ist oben mit T9 Tastaturen? Wieso tun 7 und 9 haben 4 Briefe, aber andere 3. Warum jene zwei? Warum nicht zuweisen 1 einige Briefe? Ist 0 immer Weltraum gewesen? | **C:** बिस्तर समय कहानी समय! T9 कीबोर्ड के ऊपर क्या है? क्यों 7 और 9 के पास 4 खत हैं, लेकिन अन्य 3. क्यों वो दो? क्यों नहीं 1 को कुछ खत देते? क्या 0 हमेशा अंतरिक्ष रहा है? |

Table 5: Example intentionally poor translations based on the source. This way, A > B > C. Each evaluation screen contained a document with three sources and three translations (based on one of the "models").

## E Server Speed

Part of Pearmut's flexibility comes from not using a static database to store annotation data. We test whether this has performance implications in contrast to Appraise, which uses Django templates and SQLite database. We use a laptop with i7-1260P running Ubuntu 25.10 after completing annotations in Section 5. This device was also used as a server for the annotations with Pearmut and Appraise running and accessible to the Internet via a tunnel. Each request is repeated 100 times and results are shown in Table 7.

| | **Appraise** local | **Pearmut** local | **Appraise** Internet | **Pearmut** Internet |
|---|---|---|---|---|
| Adding a campaign | 3038.7$_{\pm 65.5}$ ms | 139.0$_{\pm 8.4}$ ms | N/A | 24.0$_{\pm 1.0}$ ms+ 107.7$_{\pm 6.7}$ ms |
| Annotation screen | 127.4$_{\pm 1.2}$ ms | 1.0$_{\pm 0.1}$ ms+ 0.8$_{\pm 0.0}$ ms | 155.9$_{\pm 3.5}$ ms | 25.5$_{\pm 3.7}$ ms+ 25.5$_{\pm 1.8}$ ms |
| Campaign overview | 80.5$_{\pm 1.2}$ ms | 0.7$_{\pm 0.0}$ ms+ 0.8$_{\pm 0.0}$ ms | 105.5$_{\pm 4.3}$ ms | 24.0$_{\pm 1.0}$ ms+ 27.5$_{\pm 2.7}$ ms |
| Intermediate model ranking | N/A | 0.9$_{\pm 0.0}$ ms | N/A | 28.7$_{\pm 3.5}$ ms |
| Annotation download | 6.8$_{\pm 0.3}$ ms | 1.4$_{\pm 0.0}$ ms | N/A | 30.0$_{\pm 3.5}$ ms |

Table 7: Average time to response measured either on a local loop or via Internet connection. The ± shows the range of a 99% t-test confidence interval across 100 runs. Numbers for Pearmut are occasionally gray to distinguish between loading the website (happens once) and making a standalone request. This does not apply to Appraise which always renders templates with the data. Network benchmarks ("Internet") were conducted over a standard university WiFi connection (approx. 50Mbps, 10ms ping) to simulate realistic crowd-worker conditions.

**Scalability.** Together with the transfer, assuming single-thread execution with no concurrent parallelism and always worst-case cold-reload, a single user request takes 50ms (based on real measurements in Table 7). We assume an average annotation time per item to be 130s based on Section 5. Simply computing 130s/50ms leads to 2600, which is the maximum sustainable throughput under perfect timing.

For a more realistic computation, we define the Service-Level Agreement (performance goal) to be: "99% of users receive a response within 1000ms." The 1000ms is chosen because, despite being noticeable, it does not incur too much waiting on the annotators' part.

We model the server using an M/M/1 queue (single-thread). The service rate is $\mu = 1 / 0.050 = 20$ req/s and users send requests every 130s, giving a per-user arrival rate $\lambda_{\text{user}} = 1/130$ req/s. The cumulative distribution function for response time $T$ is based on the Poisson process (Kleinrock, 1975):

$$P(T \leq t) = 1 - e^{-(\mu-\lambda)t}$$

with SLA target of $t = 1$s. We set the probability to 0.99 and solve for the total system arrival rate $\lambda$:

$$\begin{aligned} 1 - e^{-(\mu-\lambda)\cdot 1.0} &= 0.99 \\ e^{-(\mu-\lambda)} &= 0.01 \\ -(\mu - \lambda) &= \ln(0.01) \\ \lambda &= \mu + \ln(0.01) \end{aligned}$$

Substituting the known values:

$$\lambda \approx 20.0 - 4.61 \approx 15.39 \text{ req/s}$$

Finally, we calculate the maximum number of parallel users $N$:

$$N = \frac{\lambda}{\lambda_{\text{user}}} = \frac{15.39}{\frac{1}{130}} \approx 2001$$

| | **Appraise** | **Pearmut** |
|---|---|---|
| Global | 0.792 | 0.774 |
| Group-by-model | 0.363 | 0.526 |
| Group-by-item | 0.860 | 0.697 |

Table 6: Inter-annotator agreement for three English→Czech annotators (averaged) using Pearson for Global and Group-by-model and Kendall$_b$ correlation for group-by-item.

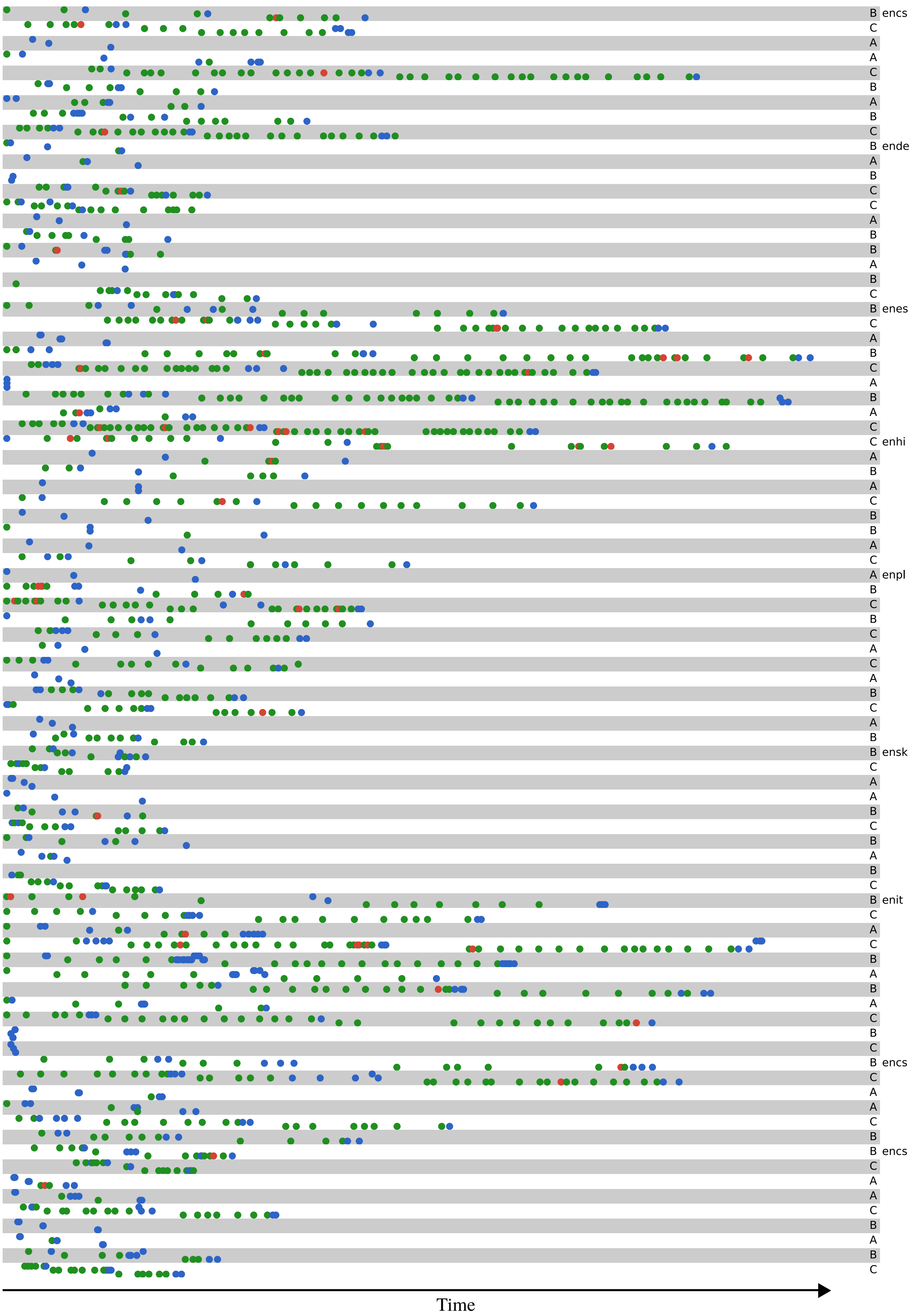

Figure 6: Time diagram of actions based on user annotations. The ● marks creating a new error span, ● deleting it, and ● assigning a score. Each document consists of three segments (one horizontal block in figure) and the correspondence of action to the segment is marked using the x-axis within that block.

The more realistic 2000 parallel users is still well above the 158 annotators that worked on a large-scale human evaluation campaign at WMT 2025 (Kocmi et al., 2025) over the period of two months (i.e. even lower number of *concurrent* users). To further scale up, it is possible to either run Pearmut on a server-grade compute, or to launch different Pearmut campaigns (e.g. one per language pair as in WMT) on different machines.

# F Example Campaign Definitions

Pearmut can be launched in three steps:

```
pip install pearmut
pearmut add my_campaign.json
pearmut run
```

## F.1 Simple ESA annotation

The simplest annotation scheme is where all the annotators annotate from the same pool of evaluation items, so we only need to define the list of items and the number of user links we wish to generate.

```
{
  "info": {
    "assignment": "single-stream",
    // DA: scores
    // ESA: error spans and scores
    // MQM: error spans, categories, and scores
    "protocol": "ESA",
    "users": 10,
  },
  "campaign_id": "my_first_campaign",
  "data": [
    // each item is a document = list
    [
      // each evaluation item is a document
      {
        // message to show to users above the first item
        "instructions": "Evaluate translation from English to Czech",
        "src": "This will be the year that Guinness loses its cool. Cheers to that!",
        "tgt": {"modelA": "Nevím přesně, kdy jsem to poprvé zaznamenal. Možná to bylo ve chvíli, ..."}
      },
      {
        "src": "I'm not sure I can remember exactly when I sensed it. Maybe it was when some...",
        "tgt": {"modelA": "Tohle bude rok, kdy Guinness přijde o svůj „cool“ faktor. Na zdraví!"}
      }
      ...
    ],
    // more documents
    ...
  ]
}
```

## F.2 Task-based contrastive annotation

In task-based assignment, the exact order of evaluation items is pre-defined by the practitioner managing the campaign. contrastive evaluation is automatically turned on when multiple model keys are specified in a document. By default, the model ordering for each document is shuffled to avoid positional bias, though this can be turned off with `info>shuffle: false`.

```
{
  "info": {
    "assignment": "task-based",
    "protocol": "ESA",
  },
  "campaign_id": "my_second_campaign",
  "data": [
    // data for first task/user
    [
      [
        // first document for first user
        {
          "src": "This will be the year that Guinness loses its cool. Cheers to that!",
          "tgt": {
            "modelA": "Tohle bude rok, kdy Guinness přijde o svůj „cool“ faktor. Na zdraví!",
            "modelB": "Toto bude rok kdy Guiness přestane být cool. Super!"
          }
        },
        {
          "src": "I'm not sure I can remember exactly when I sensed it. Maybe it was when some...",
          "tgt": {
```

```
            "modelA": "Nevím přesně, kdy jsem to poprvé zaznamenal. Možná to bylo ve chvíli, ...",
            "modelB": "Nejsem si jistý kdy jsem to zaznamenal. Možná to bylo když ..."
          }
        }
        ...
      ],
      [
        // another document for first user
        {
          "src": "I'm walking in the streets tonight, the direction to your house.",
          "tgt": {
            "modelC": "Chodním dnes ulicemi, směrem k tobě domů.",
            "modelB": "Dnes večer jdu ulicemi, směrem k tvému domu."
          }
        },
        ...
      ],
      // more documents
      ...
    ],
    // data for second task/user
    [
        ...
    ],
    // arbitrary number of users (each corresponds to a single URL to be shared)
  ]
}
```

### F.3 Tutorial and attention checks

Tutorials and attention checks are the same as other evaluation items but with the `validation` key. Tutorial itself is simply an attention check with a warning on failure and usually with a skip button enable so that an annotator that already went through the tutorial can skip it.

```
{
  "src": "The quick brown fox jumps.",
  "tgt": {"modelA": "Rychlá hnědá liška skáče."},
  "validation": {
    "modelA": [
      {
        "warning": "Please set score between 70-80.",  // shown on failure (omit for silent logging)
        "score": [70, 80],                             // required score range [min, max]
        "error_spans": [{"start_i": [0, 2], "end_i": [4, 8], "severity": "minor"}],  // expected spans
        "allow_skip": true                             // show "skip tutorial" button
      }
    ]
  }
}
```

For contrastive evaluation, a special validation is available:

```
{
  "src": "AI transforms industries.",
  "tgt": {"A": "UI transformuje průmysly.", "B": "Umělá inteligence mění obory."},
  "validation": {
    "A": [
      {"warning": "A has error, score 20-40.", "score": [20, 40]}
    ],
    "B": [
      {"warning": "B must score higher than A.", "score": [70, 90], "score_greaterthan": "A"}
    ]
  }
}
```

## G Usage of Human Evaluation in Papers

We selected all NAACL 2025, ACL 2025, and EMNLP 2025 papers, including findings that contain the word *Translation* or *MT* in their titles. We intentionally did not include WMT and IWSLT papers as these occasionally rely on human evaluation from a shared task. This yielded 108 papers, which were then manually classified into categories: *human evaluation is present*, including small scale (26, 24%), *human evaluation is missing* (55, 51%), and *human evaluation is not required* (27, 25%). When the human evaluation takes place, it is oftentimes with an ad-hoc annotation protocol or not described at all, which makes it irreproducible and unauditable. The complete list follows:

- ✅ Understanding In-Context Machine Translation for Low-Resource Languages: A Case Study on Manchu, DA
- ✅ Locate-and-Focus: Enhancing Terminology Translation in Speech Language Models, terminology verification
- ✅ Lost in Literalism: How Supervised Training Shapes Translationese in LLMs, translationese span ratio + ranking
- ✅ SwiLTra-Bench: The Swiss Legal Translation Benchmark, DA
- ✅ It's Not a Walk in the Park! Challenges of Idiom Translation in Speech-to-text Systems, error categorization
- ✅ Towards Style Alignment in Cross-Cultural Translation, pairwise
- ✅ SHIFT: Selected Helpful Informative Frame for Video-guided Machine Translation, pairwise
- ✅ A Case Against Implicit Standards: Homophone Normalization in Machine Translation for Languages that use the Ge'ez Script. ⭐, qualitative analysis + ranking
- ✅ Viability of Machine Translation for Healthcare in Low-Resourced Languages, adequacy + clinical risk
- ✅ MultiMed-ST: Large-scale Many-to-many Multilingual Medical Speech Translation, adequacy, fluency, comprehensibility (1 to 10)
- ✅ Should I Share this Translation? Evaluating Quality Feedback for User Reliance on Machine Translation, would-share decision
- ✅ MAVL: A Multilingual Audio-Video Lyrics Dataset for Animated Song Translation, singability, sense, overall (1 to 5)
- ✅ Languages Still Left Behind: Toward a Better Multilingual Machine Translation Benchmark, MQM
- ✅ Translate Smart, not Hard: Cascaded Translation Systems with Quality-Aware Deferral, DA
- ✅ Leveraging Loanword Constraints for Improving Machine Translation in a Low-Resource Multilingual Context, DA
- ✅ LiTransProQA: An LLM-based Literary Translation Evaluation Metric with Professional Question Answering, MQM + pairwise
- ✅ Toward Machine Translation Literacy: How Lay Users Perceive and Rely on Imperfect Translations, task-based
- ✅ Liaozhai through the Looking-Glass: On Paratextual Explicitation of Culture-Bound Terms in Machine Translation, pairwise
- ✅ Towards Zero-Shot Multimodal Machine Translation, pairwise
- ✅ How to Learn in a Noisy World? Self-Correcting the Real-World Data Noise in Machine Translation, adequacy (1 to 5)
- ✅ Investigating Hallucinations in Simultaneous Machine Translation: Knowledge Distillation Solution and Components Analysis, hallucination analysis
- ✅ Large Language Models for Persian-English Idiom Translation, fluency (1 to 5) + idiom (0 or 1)
- ✅ Automatic Input Rewriting Improves Translation with Large Language Models, fluency, understandability, meaning (1 to 4)
- ✅ How Good Are LLMs for Literary Translation, Really? Literary Translation Evaluation with Humans and LLMs, MQM + DA
- ✅ Scaling Low-Resource MT via Synthetic Data Generation with LLMs, DA
- ✅ AFRIDOC-MT: Document-level MT Corpus for African Languages, DA + ESA
- ❌ DMDTEval: An Evaluation and Analysis of LLMs on Disambiguation in Multi-domain Translation
- ❌ Towards Building Large Scale Datasets and State-of-the-Art Automatic Speech Translation Systems for 14 Indian Languages
- ❌ Single-to-mix Modality Alignment with Multimodal Large Language Model for Document Image Machine Translation
- ❌ Did Translation Models Get More Robust Without Anyone Even Noticing? ⭐
- ❌ Unveiling the Power of Source: Source-based Minimum Bayes Risk Decoding for Neural Machine Translation
- ❌ Alleviating Distribution Shift in Synthetic Data for Machine Translation Quality Estimation
- ❌ Translation and Fusion Improves Cross-lingual Information Extraction
- ❌ Enhancing Neural Machine Translation Through Target Language Data: A $k$NN-LM Approach for Domain Adaptation
- ❌ Two Intermediate Translations Are Better Than One: Fine-tuning LLMs for Document-level Translation Refinement
- ❌ MLAS-LoRA: Language-Aware Parameters Detection and LoRA-Based Knowledge Transfer for Multilingual Machine Translation
- ❌ SimulS2S-LLM: Unlocking Simultaneous Inference of Speech LLMs for Speech-to-Speech Translation
- ❌ Improving Language and Modality Transfer in Translation by Character-level Modeling

- ❌ THOR-MoE: Hierarchical Task-Guided and Context-Responsive Routing for Neural Machine Translation
- ❌ Registering Source Tokens to Target Language Spaces in Multilingual Neural Machine Translation
- ❌ Enhancing Machine Translation with Self-Supervised Preference Data
- ❌ Make Imagination Clearer! Stable Diffusion-based Visual Imagination for Multimodal Machine Translation
- ❌ Multi-perspective Alignment for Increasing Naturalness in Neural Machine Translation ⭐
- ❌ An Empirical Study of Iterative Refinements for Non-autoregressive Translation
- ❌ GrammaMT: Improving Machine Translation with Grammar-Informed In-Context Learning
- ❌ SLoW: Select Low-frequency Words! Automatic Dictionary Selection for Translation on Large Language Models
- ❌ Whisper-UT: A Unified Translation Framework for Speech and Text
- ❌ Few-Shot Learning Translation from New Languages
- ❌ PRIM: Towards Practical In-Image Multilingual Machine Translation
- ❌ PoseStitch-SLT: Linguistically Inspired Pose-Stitching for End-to-End Sign Language Translation
- ❌ Please Translate Again: Two Simple Experiments on Whether Human-Like Reasoning Helps Translation
- ❌ EnAnchored-X2X: English-Anchored Optimization for Many-to-Many Translation
- ❌ Dynamic Jointly Batch Selection for Data Efficient Machine Translation Fine-Tuning
- ❌ You Are What You Train: Effects of Data Composition on Training Context-aware Machine Translation Models
- ❌ BOUQuET : dataset, Benchmark and Open initiative for Universal Quality Evaluation in Translation
- ❌ DrFrattn: Directly Learn Adaptive Policy from Attention for Simultaneous Machine Translation
- ❌ In-Context Example Selection via Similarity Search Improves Low-Resource Machine Translation
- ❌ Beyond English: The Impact of Prompt Translation Strategies across Languages and Tasks in Multilingual LLMs
- ❌ TEaR: Improving LLM-based Machine Translation with Systematic Self-Refinement
- ❌ Effective Self-Mining of In-Context Examples for Unsupervised Machine Translation with LLMs
- ❌ DOLFIN - Document-Level Financial Test-Set for Machine Translation
- ❌ Dynamic Feature Fusion for Sign Language Translation Using HyperNetworks
- ❌ Beyond the Mode: Sequence-Level Distillation of Multilingual Translation Models for Low-Resource Language Pairs
- ❌ Media of Langue: Exploring Word Translation Network
- ❌ CA*: Addressing Evaluation Pitfalls in Computation-Aware Latency for Simultaneous Speech Translation
- ❌ LITERA: An LLM Based Approach to Latin-to-English Translation
- ❌ Faster Machine Translation Ensembling with Reinforcement Learning and Competitive Correction
- ❌ MoCE: Adaptive Mixture of Contextualization Experts for Byte-based Neural Machine Translation
- ❌ Detect, Disambiguate, and Translate: On-Demand Visual Reasoning for Multimodal Machine Translation with Large Vision-Language Models
- ❌ The Impact of Domain-Specific Terminology on Machine Translation for Finance in European Languages ⭐
- ❌ A Bayesian Optimization Approach to Machine Translation Reranking
- ❌ Mitigating Hallucinated Translations in Large Language Models with Hallucination-focused Preference Optimization
- ❌ An Efficient Gloss-Free Sign Language Translation Using Spatial Configurations and Motion Dynamics with LLMs
- ❌ How to Align Multiple Signed Language Corpora for Better Sign-to-Sign Translations? ⭐
- ❌ Multilingual Machine Translation with Open Large Language Models at Practical Scale: An Empirical Study
- ❌ Anticipating Future with Large Language Model for Simultaneous Machine Translation
- ❌ Towards Inducing Long-Context Abilities in Multilingual Neural Machine Translation Models
- ❌ Is Translation All You Need? A Study on Solving Multilingual Tasks with Large Language Models
- ❌ Continual Learning in Multilingual Sign Language Translation
- ❌ Gender Bias in Nepali-English Machine Translation: A Comparison of LLMs and Existing MT Systems
- ❌ ConECT Dataset: Overcoming Data Scarcity in Context-Aware E-Commerce MT
- ⚪ Extending Automatic Machine Translation Evaluation to Book-Length Documents
- ⚪ Assumed Identities: Quantifying Gender Bias in Machine Translation of Gender-Ambiguous Occupational Terms
- ⚪ Mind the Inclusivity Gap: Multilingual Gender-Neutral Translation Evaluation with mGeNTE
- ⚪ M-MAD: Multidimensional Multi-Agent Debate for Advanced Machine Translation Evaluation
- ⚪ Adaptive and Robust Translation from Natural Language to Multi-model Query Languages
- ⚪ Enhancing Human Evaluation in Machine Translation with Comparative Judgement
- ⚪ Watching the Watchers: Exposing Gender Disparities in Machine Translation Quality Estimation
- ⚪ Magnet: Multi-turn Tool-use Data Synthesis and Distillation via Graph Translation
- ⚪ ReMedy: Learning Machine Translation Evaluation from Human Preferences with Reward Modeling
- ⚪ ExeCoder: Empowering Large Language Models with Executability Representation for Code Translation
- ⚪ Same evaluation, more tokens: On the effect of input length for machine translation evaluation using Large Language Models
- ⚪ Evaluating Language Translation Models by Playing Telephone
- ⚪ From Tens of Hours to Tens of Thousands: Scaling Back-Translation for Speech Recognition

- ⚪ Translationese-index: Using Likelihood Ratios for Graded and Generalizable Measurement of Translationese
- ⚪ Translation in the Hands of Many: Centering Lay Users in Machine Translation Interactions
- ⚪ Unsupervised Word-level Quality Estimation for Machine Translation Through the Lens of Annotators (Dis)agreement
- ⚪ Multimodal Neural Machine Translation: A Survey of the State of the Art
- ⚪ An Interdisciplinary Approach to Human-Centered Machine Translation
- ⚪ Explicit Learning and the LLM in Machine Translation
- ⚪ Automatic Annotation Augmentation Boosts Translation between Molecules and Natural Language
- ⚪ Fingerspelling within Sign Language Translation
- ⚪ MSc-SQL: Multi-Sample Critiquing Small Language Models For Text-To-SQL Translation
- ⚪ AI-Assisted Human Evaluation of Machine Translation
- ⚪ Few-Shot Natural Language to First-Order Logic Translation via Code Generation
- ⚪ LLM-Supported Natural Language to Bash Translation
- ⚪ SSA-COMET: Do LLMs Outperform Learned Metrics in Evaluating MT for Under-Resourced African Languages?
- ⚪ PromptOptMe: Error-Aware Prompt Compression for LLM-based MT Evaluation Metrics

Note that the categorization is not a criticism of the works but an observation of missing human evaluation. In fact, many papers openly admit the lack of human evaluation in their Limitation sections (marked with ⭐). This study is similar to that of Marie et al. (2021) with at most 15% of papers using human evaluation. Our analysis is on a smaller scale and with a more conservative criterion of classifying a paper as lacking human evaluation, and with filtering out papers that might not need any human evaluation. We attribute the general lack of human evaluation partially to the high bar of entry to set up an evaluation tool but also to a high trust in automated machine translation metrics, which might be, unfortunately, misplaced (Lavie et al., 2025; Kocmi et al., 2024b).